\documentclass[sigconf, authordraft, review=false]{acmart} 

\renewcommand\footnotetextcopyrightpermission[1]{} 
\usepackage{adjustbox}

\usepackage{microtype}

\usepackage{hyperref}

\usepackage{paralist}

\usepackage{tikz}
\usepackage{tikz-qtree}
\usetikzlibrary{calc,trees,positioning,arrows,chains,shapes.geometric,%
    decorations.pathreplacing,decorations.pathmorphing, decorations.markings, shapes,%
    matrix,shapes.symbols,scopes}

\newcommand{\nonterm}[1]{{\color{blue} #1}}
\newcommand{\leaf}[1]{{\color{red} #1}} 	

\usepackage{booktabs}
\usepackage{tabularx}
\usepackage{multirow}

\usepackage{array}

\usepackage{amsmath}
\usepackage{amssymb}

\usepackage{graphicx}
\graphicspath{ {images/} }

\usepackage{syntax}

\usepackage{url}
\urldef{\mailsa}\path|{kevin, psk}@cse.iitm.ac.in|

\usepackage{kevinmacros}

\usepackage{booktabs} 

\setcopyright{rightsretained}

\acmDOI{10.475/123_4}

\acmISBN{123-4567-24-567/08/06}

\acmConference[K-CAP]{ACM Woodstock conference}{December 2017}{Austin, Texas, United States} 
\acmYear{1997}
\copyrightyear{2016}

\acmPrice{15.00}

\acmSubmissionID{}

\begin{document}\sloppy
\title[Extracting Ontological Knowledge from Textual Descriptions]{Extracting Ontological Knowledge from Textual Descriptions}

\author{Kevin Alex Mathews}
\affiliation{%
  \institution{Indian Institute of Technology Madras}}
\email{kevin@cse.iitm.ac.in}

\author{P Sreenivasa Kumar}
\affiliation{%
  \institution{Indian Institute of Technology Madras}
  }
\email{psk@cse.iitm.ac.in}

\renewcommand{\shortauthors}{Kevin Alex Mathews and P Sreenivasa Kumar}


\begin{abstract}
%
%

Authoring of \owldl ontologies is intellectually challenging and to make this process simpler, many systems accept natural language text as input.
A text-based ontology authoring approach can be successful only when it is combined with an effective method for extracting ontological axioms from text.
Extracting axioms from unrestricted English input is a substantially challenging task due to the richness of the language.
Controlled natural languages (\cnls) have been proposed in this context and these tend to be highly restrictive.
In this paper, we propose a new \cnl called \theGrammar (TExtual DEscription Identifier) whose grammar is inspired by the different ways OWL-DL constructs are expressed in English.
We built a system that transforms \theGrammar sentences into corresponding \owldl axioms.
Now, ambiguity due to different possible lexicalizations of sentences and semantic ambiguity present in sentences are challenges in this context.
We find that the best way to handle these challenges is to construct axioms corresponding to alternative formalizations of the sentence so that the end-user can make an appropriate choice.
The output is compared against human-authored axioms and in substantial number of cases, human-authored axiom is indeed one of the alternatives given by the system.
The proposed system substantially enhances the types of sentence structures that can be used for ontology authoring.

\end{abstract}

%
%

\keywords{Text Analytics, Ontology Learning, Knowledge Extraction} 

\maketitle
\section{Introduction}
\label{section:introduction}


The key to building large and powerful AI systems is knowledge representation. An ontology is a knowledge representation mechanism. It provides a vocabulary describing a domain of interest, and a specification of the terms in that vocabulary.
However the manual creation of ontologies is intellectually challenging and time-consuming\mycite{Buitelaar2004}.
Also the knowledge in an ontology is generally expressed in ontology languages such as \rdfs\mycite{Brickley2004} or \owl\mycite{DeborahL.McGuinness2004}, which are based on \dl\mycite{Baader2010a}. As a result, it is difficult for non-logicians to create, edit or manage ontologies.
So a user-friendly format for communication of ontological content is required for ontology authoring.
Many systems accept natural language (English) text as input.
Now a text-based ontology authoring system can be successful only when it is combined with an effective method for extracting ontological axioms from text.
Extracting axioms from English is challenging due to its unrestricted and ambiguous nature.
\cnls, which are restricted unambiguous variants of natural languages, have been proposed in this context.

Now,  ambiguity due to different possible lexicalizations of sentences and semantic ambiguity present in sentences are challenges in the context of ontology authoring.
It is difficult to disambiguate the input sentence without substantial background knowledge of the domain.
Most current authoring systems produce one formalization of a sentence.
We find that the best way to handle these challenges is to construct axioms corresponding to alternative formalizations of the sentence so that the end-user can make an appropriate choice.

In this paper, we propose a novel ontology authoring system that simplifies the authoring process for the users.
Our system takes English text that conforms to a newly proposed \cnl called \theGrammar input and generates corresponding \owl axioms.
We scope ontology authoring to building the schema (\emph{TBox}) of an ontology. 
Within this context, we focus on \owl class expression axioms.


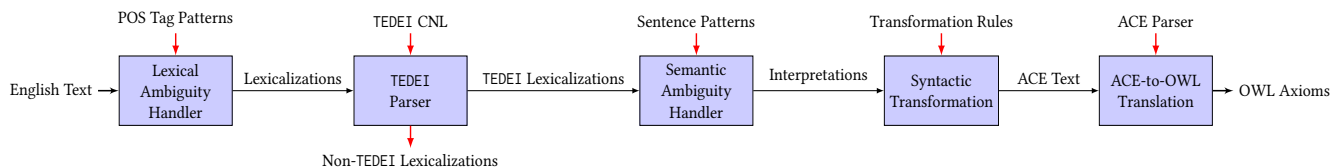
\begin{figure*}
\centering
\tikzstyle{myblock} = [rectangle, minimum height=4em, minimum width=7em]
\tikzstyle{block} = [draw, fill=blue!20, rectangle, minimum height=4em, minimum width=7em]
\tikzstyle{pinstyle} = [pin edge={latex-,thick,red}]
\tikzstyle{revpinstyle} = [pin edge={-latex,thick,red}]
\tikzstyle{mastyle} = [pin edge={<-,thick, black}]
\tikzstyle{revmastyle} = [pin edge={->,thick,black}]

\resizebox{\textwidth}{7em}
{
\begin{tikzpicture}[auto, >=latex']	
\node [block, node distance=3cm, pin={[pinstyle]above:POS Tag Patterns}, pin={[mastyle]left:English Text}, align=center] (la) {Lexical \\Ambiguity \\Handler};
\node [block, right of=la, node distance=4.6cm, pin={[pinstyle]above:\theGrammar \cnl}, , pin={[revpinstyle]below:Non-\theGrammar Lexicalizations}, align=center] (tp) { \theGrammar \\ Parser};
\node [block, right of=tp, node distance=5.6cm, pin={[pinstyle]above:Sentence Patterns}, align=center] (sa) {Semantic \\Ambiguity \\Handler};
\node [block, right of=sa, node distance=4.8cm, pin={[pinstyle]above:Transformation Rules}, align=center] (st) {Syntactic \\ Transformation};
\node [block, right of=st, node distance=4.2cm, pin={[pinstyle]above:\ace Parser}, pin={[revmastyle]right:\owl Axioms}, align=center] (aot) {\ace-to-\owl \\ Translation};

\draw [-latex] (la)  -- node {Lexicalizations} (tp);
\draw [-latex] (tp) -- node [above] {\theGrammar Lexicalizations} (sa);
\draw [-latex] (sa) -- node {Interpretations} (st);
\draw [-latex] (st)  -- node {\ace Text} (aot);
\end{tikzpicture}
}
\caption{Architecture}
\label{fig:arch-pipe}
\end{figure*}

We have outlined the architecture of our system in \figref{fig:arch-pipe}.
The five main modules of the system are lexical ambiguity handler, \theGrammar parser, semantic ambiguity handler, syntactic transformation and \ace-to-\owl translation.
Lexical ambiguity handler accepts the input sentences, expressed in English, and generates possible lexicalizations of the sentence using POS tag patterns.
Lexicalization is the process of breaking the given sentence into tokens
	such that these tokens can be identified as various ontology elements, namely, classes, individuals, properties and concept constructors.
Then \theGrammar parser parses the lexicalizations on the basis of \theGrammar grammar rules.
Only valid \theGrammar lexicalizations are processed further.
Sentences not conforming to \theGrammar are indicated as such to the user so that they can be reformulated.
Semantic ambiguity handler generates possible interpretations of the \theGrammar lexicalizations using specific sentence patterns.
The interpretations are converted to \ace in the syntactic transformation module using rules of transformation.
Finally \ace sentences are converted to \owl by the existing \ace parser.

The output of the system is compared against human-authored axioms and in substantial number of cases, human-authored axiom is indeed one of the alternatives given by the system.
Our framework clearly outperforms \ace in terms of the number and types of sentences the system can handle.
In comparison with existing systems, due to the use of \theGrammar, our framework is a robust way to generate ontological axioms from text.
\theGrammar helps in clearly defining the scope of the system and provides the ability to reject a sentence and ask for reformulation.
Also, employing \ace as an intermediate language aids formalization and reduces the complexity of the system.


Our contributions in this paper are as follows: 
(1) an ontology authoring process,
(2) an ontology authoring language whose grammar reflects the constructs of OWL and which has better expressivity than ACE,
(3) extraction of OWL axioms from sentences of the language, and
(4) handling ambiguity associated with formalization.


The remainder of the paper is structured as follows: 
in \secref{section:related-works}, we discuss the related works.
In \secref{section:tedei}, we discuss the important grammar rules of \theGrammar.
In \secref{section:ace}, we discuss \ace and transformation of \theGrammar text to \ace.
In \secref{section:ambiguity}, we describe ambiguity in formalization and how it is handled by the proposed system.
\secref{section:results} describes the results and evaluation and \secref{section:conclusion} concludes the paper.

\section{Related Works}\label{section:related-works}

In this section, we discuss the works related to ontology authoring and 
	how the proposed system is an improvement over the existing systems.
\secref{section:cnl-authoring} discusses ontology authoring based on CNLs.
Note that a text-based ontology authoring approach needs to have a corresponding solution to the problem of extracting ontology from text. 
The success of the authoring tool depends on an effective technique for extraction. 
Extracting ontology is also termed as ontology learning in the literature and we review some of the relevant works in \secref{section:auto-authoring}.

\subsection{CNL-based ontology authoring}\label{section:cnl-authoring}
Some \cnls that were developed specifically 
	for creating \owl ontologies are \ace\mycite{Fuchs2008}, Rabbit\mycite{dolbear2007rabbit}, CLOnE\mycite{Funk2007} and SOS\mycite{Cregan2007}. 
%
However \ace is the most commonly used \cnl for Semantic Web.
There are numerous tools based on \ace such as ACEWiki\footnote{http://attempto.ifi.uzh.ch/acewiki/} and APE\footnote{http://attempto.ifi.uzh.ch/ape/}.
\ace is designed to be unambiguous and less complex as compared to standard English. 
In the context of ontologies, \ace offers a simple platform for domain experts who are not comfortable with ontology languages to author ontologies.
ACE-OWL, a sublanguage of \ace, has a bidirectional mapping with \owl.

Although \ace is used for ontology authoring, \ace has two main limitations.
Firstly, \ace is a subset of standard English.
As a result, many English sentences are not present in \ace. 
Secondly, in some cases, even though a sentence is valid in \ace, 
	the \ace parser, named Attempto Parsing Engine (\ape), fails to axiomatize it because only a subset of \ace sentences map to \owl axioms.
Thus we find that though \ace is good for authoring ontologies, it restricts the user to a limited subset of English. 

However, we noted that with suitable transformation, it is possible to make sentences \ace-compliant or \owl-compliant. 
The proposed system addresses both the limitations of \ace in the syntactic transformation phase.
Syntactic transformation carries out the necessary transformation and 
	as a result, it extends \ace for the purpose of ontology authoring.

\subsection{Ontology learning}\label{section:auto-authoring}


The main techniques in ontology learning from text are linguistics-based and statistics-based. 
Linguistics-based techniques involve exploitation of lexico-syntactic patterns\mycite{hearst1992automatic}, utilization of knowledge-rich resources such as linked data or ontologies\mycite{Gangemi2016a}, and syntactic transformation\mycite{Volker2007a}.
Statistics-based techniques involve relevance analysis\mycite{Buhmann2014a}, co-occurrence analysis, clustering, formal concept analysis, association rule mining\mycite{Drymonas2010a} and deep learning\mycite{Petrucci2016a}.
Wong et al.\mycite{Wong2012a} presents a study on prominent ontology learning systems such as OntoLearn Reloaded\mycite{Velardi2013}, Text2Onto\mycite{CimianoPhilippandVolker2005}, 
and OntoGain\mycite{Drymonas2010a}. Most of the current systems employ a combination of the aforementioned techniques.

In comparison with existing systems, the proposed system has a streamlined approach to the generation of axioms from text. 
Existing systems attempt to convert any English sentence and the process would either fail or generate an incorrect axiom in several cases. 
However, in the proposed approach, since it is guided by a grammar, the system can identify a sentence it can not handle and give an error signal.
Although existing systems might be using a grammar, since it is not explicitly mentioned, it is difficult to define the scope of the input.
In our framework, the use of grammar clearly defines the set of sentences the system can handle.
Our approach gives the end-user an opportunity to rewrite the sentence and possibly convey the same information in a different way.

In addition, by using \ace in our framework, we incorporate the advantage of using a \cnl in our framework.
Also, existing systems generate only one formalization per sentence without taking into the account the impact of ambiguity in formalization.




\section{Text Description Identifier (\theGrammar)}\label{section:tedei}

Only sentences having at least one lexicalization that is valid according to \theGrammar rules are handled by our system.
The expressivity of DL covered by the language defined by \theGrammar is ALCQ. 
We use \antlr\mycite{Parr1995} parser generator to generate the parser for \theGrammar.
\antlr takes \theGrammar grammar as input and generates a recognizer for the language.
We employ the recognizer to read the input stream and check whether it conforms to the syntax specified by \theGrammar.

We developed \theGrammar keeping in mind the rich set of primitives of \owltwo\mycite{Grau2008}---the W3C recommended and widely adopted ontology language.
We also mined Brown corpus\mycite{francis1979brown} to identify various ways in which \owl primitives can be expressed in natural language.
We used various regular expressions to extract the patterns.
Brown corpus, being a prominent text corpus, is rich enough to contain various kinds of such patterns.
Those patterns that correspond to some OWL primitive are encoded as rules in the grammar.


The non-terminals of the grammar correspond to \owl primitives. In this section, we describe only the important rules of the grammar.
The primary rule in \theGrammar corresponds to a class expression axiom. A class expression axiom defines a concept. The textual description corresponding to a class expression axiom consists of a concept and a class expression. An example is the sentence \emph{``Every square contains right angles,''} which consists of the concept \emph{square} and the class expression \emph{contains right angles}.

A class expression can either be an atomic class or a complex class.
An atomic class denotes a single concept.
A complex class can be constructed in various ways.
It is possible to employ set operations (union, intersection, and complement) and property restrictions (existential restriction, universal restriction, cardinality restriction, value restriction, and self-restriction.)
It is also possible to construct complex classes by enumerating all the instances in the class.

The natural language indicators for various \owl primitives are given in the production rules of the grammar.
Brown corpus has been beneficial in identifying these indicators.
We consider the relative pronouns 
\emph{which}, \emph{that}, and \emph{who} as the natural language indicators for intersection of classes. An example is the sentence \emph{``Every square is a quadrilateral that has 4 right angles.''} 
Sometimes, the absence of any such relative pronoun also indicates an intersection, as demonstrated by the sentence \emph{``Every rectangle is a quadrilateral having 4 right angles.''} The natural language indicator for union of classes is the conjunction \emph{or}.
An example is the sentence \emph{``Every polygon is concave or convex.''} The determiner \emph{some}
indicates existential restriction. An example is the sentence \emph{``Every rectangle contains some right angles.''} The determiner \emph{only} 
indicates universal restriction. An example is the sentence \emph{``Every rectangle contains only right angles.''}

The important rules of \theGrammar are outlined below in BNF notation. For the sake of readability, we provide only an abstract version of the rules originally written in \antlr. The non-terminals of the grammar are shown in italics and terminals are shown in uppercase.
\texttt{CLASS}, \texttt{INDIVIDUAL}, and \texttt{PROPERTY} are three special terminals of the grammar, 
	and they correspond to ontology elements class, individual and property respectively.
The actual text tokens that correspond to these terminals are identified during the lexicalization process.
Sometimes, there may be multiple lexicalizations for a sentence and 
	we discuss the details of the lexicalization process in \secref{section:lex-amb}.
Rest of the terminals such as \texttt{THAT} and \texttt{WHICH} correspond to words or phrases of the input sentence.

\setlength{\grammarparsep}{4pt plus 0.5pt minus 0.5pt}
\setlength{\grammarindent}{3.9em}

\renewcommand{\syntleft}{\normalfont\itshape}
\renewcommand{\syntright}{}

\renewcommand{\litleft}{\bgroup\ulitleft\small}
\renewcommand{\litright}{\ulitright\egroup}

\def\alty{\textbar}

\begin{grammar}

<start>
	::= <lexpr> <rexpr>
      
<lexpr> ::= 
	`CLASS'
	\alty\ `INDIVIDUAL'

<rexpr>
	::= <union>


<union>
	::= (<intersection>      
	\alty\ <complement>) (<unionInd> <union>)*    

<intersection>
	::= ~\hspace{-3pt}\hbox{<clsExpComb> (<intersectionInd> <intersection>)*}

<clsExpComb>
    ::= <clsExp> (<clsExpInd>? <clsExpComb>)*


<clsExp>
	::=  <complement>   
	\alt <uniRes>        	\hspace{28pt}  	//\texttt{universal restriction}
	\alt <existlRes>     	\hspace{20pt}    	//\texttt{existential restriction} 
	\alt <exactCard>  	\hspace{14pt}	   	//\texttt{exact cardinality} 
	\alt <minCard>    	\hspace{19pt}	   	//\texttt{minimum cardinality} 
	\alt <maxCard>			\hspace{18pt}    	//\texttt{maximum cardinality}
	\alt <qualExactCard> 	\hspace{-5pt}    	//\texttt{qualified exact cardinality}
	\alt <qualMinCard>	\hspace{0pt}    	//\texttt{qualified minimum cardinality}
	\alt <qualMaxCard>	\hspace{-1pt}    	//\texttt{qualified maximum cardinality}
	\alt <indValueRes>	\hspace{2pt}    	//\texttt{individual-value restriction}
	\alt <selfValueRes>	\hspace{1pt}    	//\texttt{self-value restriction}
	\alt <classComb> 

<complement>
	::= ~\newline 
	 <preComplementInd> `PROPERTY' <classComb>
	 \alt `PROPERTY' <postComplementInd> <classComb>

<uniRes>
	::= ~\newline `PROPERTY' <universalInd> <classComb>   
	\alt  <universalInd> `PROPERTY' <classComb>   

<existRes>
	::= ~\newline `PROPERTY' <existentialInd> <classComb> 
	\alt `PROPERTY' <classComb>                   

<exactCard>
	::= ~\newline  `PROPERTY' <exactCardInd> `CARDINALITY' 
	\alt `PROPERTY' <ambiExactCardInd> `CARDINALITY'          

<minCard>
	::= ~\newline `PROPERTY' <preMinCardInd> `CARDINALITY' 
	\alt `PROPERTY' `CARDINALITY' <postMinCardInd>
	
<maxCard>
	::= ~\newline `PROPERTY' <preMaxCardInd> `CARDINALITY'
	\alt `PROPERTY' `CARDINALITY' <postMaxCardInd>

<qualExactCard>
	::= ~\newline `PROPERTY' <exactCardInd> `CARDINALITY' <classComb>
	\alt `PROPERTY' <ambiExactCard> `CARDINALITY' <classComb>         

<qualMinCard>
	::= ~\newline `PROPERTY' <preMinCardInd> `CARDINALITY' <classComb>
	\alt`PROPERTY' `CARDINALITY' <postMinCardInd> <classComb>

<qualMaxCard>
	::= ~\newline `PROPERTY' <preMaxCardInd> `CARDINALITY' <classComb>
	\alt `PROPERTY'~`CARDINALITY' <postMaxCardInd> <classComb>

<indValueRes>
	::= `PROPERTY' `INDIVIDUAL'

<selfValueRes>
	::= `PROPERTY' <selfInd>

<classComb>
    ::= `CLASS' (<clsExpInd> <classComb>)*

<clsExpInd> 					::= `AND' \alty\ `OR' \alty\ `,'            				

<unionInd> ::= `OR'

<intersectionInd> ::= `THAT' \alty\ `WHICH' \alty\ `WHO' \alty\ `WHOSE'

<preCompInd> ::= `DOES NOT' \alty\ `DO NOT' \alty\ `DID NOT'

<postCompInd> ::= `NOT' \alty\ `NO' 

<universalInd> ::= ~\newline `EXCLUSIVELY' \alty\ `NOTHING BUT' \alty\ `NOTHING EXCEPT' \alty\ `ONLY'

<existentialInd> ::= ~\newline `A' \alty\ `AN' \alty\ `ALL' \alty\ `ANY' \alty\ `FEW' \alty\ `MANY' \alty\ `SOME' \alty\ `SEVERAL'

<exactCardinalityInd> ::= `EXACTLY' \alty\ `JUST' \alty\ `MAY BE' \alty\ `ONLY'

<ambiExactCardInd> ::= ~\newline `ABOUT' \alty\ `ALMOST' \alty\ `APPROXIMATELY' \alty\ `AROUND' \alty\ `CLOSE TO'

<preMinCardInd> ::= ~\newline `ATLEAST' \alty\ `LEAST' \alty\ `MORE THAN' \alty\ `NOT LESS THAN'

<postMinCardInd> ::= `OR MORE'

<preMaxCardInd> ::= ~\newline `ATMOST' \alty\ `MOST' \alty\ `LESS THAN' \alty\ `NOT MORE THAN' \alty\ `WITHIN'

<postMaxCardInd> ::= `OR LESS'
 
<selfInd> ::= `MYSELF'   \alty\ `OURSELVES'          		     
		    				\alty\ `YOURSELF' \alty\ `YOURSELVES'                      
	       				    \alty\ `HIMSELF'  \alty\ `HERSELF'  \alty\ `ITSELF' \alty\ `THEMSELVES'  
\end{grammar}

\section{Syntactic Transformation}\label{section:ace}









The proposed system employs syntactic transformation to convert \theGrammar text (i.e. sentences having at least one \theGrammar lexicalization) to \ace text.
We implemented syntactic transformation as actions that are attached to grammar elements in \theGrammar. 
We describe here the important steps in this transformation process.

We refer to the phrases in a sentence that correspond to \owl concepts as concept-terms and the phrases that correspond to \owl relations as relation-terms. For example, in the sentence \emph{``Every adenine is a purine base found in DNA''}, \emph{purine base} is a concept-term, and \emph{found in} is a relation-term.

In a sentence, a concept-term or a relation-term may contain more than one word. 
As per the rules of \ace, multi-word terms are required to have hyphens. 
We handle such terms by inserting a hyphen between consecutive words in the term. 
For example, the concept-term \emph{purine base} is transformed to \emph{purine-base}.

If a concept-term or a relation-term is not present in the \ace lexicon, 
	we add it to the lexicon dynamically by tagging it with the prefix that indicates its word class.
All noun phrases having common nouns are identified as \owl concepts and 
	they are tagged with the prefix \emph{n}.
All verb phrases are identified as \owl relations and 
	they are tagged with the prefix \emph{v}.
All noun phrases having proper nouns are identified as \owl individuals and 
	they are tagged with the prefix \emph{p}.
For example, the concept-term \emph{purine base} is tagged with \emph{n:} to denote that it is an \owl concept.
These prefixes are as per requirement of the \ace parser.

According to the \ace construction rules, every noun should be preceded by an article. 
In the absence of an article, we insert an appropriate article. 
We consider the determiner \emph{a} to be the default article for \owl atomic concepts \todo and the determiner \emph{some} to be the default article for \owl role fillers. 

We handle missing role fillers by adding \emph{something} as the role filler. 
This transformation is semantics-preserving because \emph{something} denotes the top concept in \owl.
For example, the sentence \emph{all kids play} is transformed to \emph{all kids play something}.


In \ace, all coordinated noun phrases have to agree in number and verb form (either finite or infinite). 
In this context, noun phrases denote \owl role fillers. 
In the case of a role filler that is coordinated by \emph{and} and \emph{or}, we distribute the individual role fillers to the relation according to ACE semantics. 
For example, the coordinated noun phrase \emph{likes cats and dogs} is transformed to \emph{likes cats and likes dogs}.

In \ace, all coordinated verb phrases have to agree in number and verb form (either finite or infinite).
In this context, verb phrases denote \owl relations.
In the case of a relation that is coordinated by \emph{and} and \emph{or}, we distribute the individual relations to the role filler according to ACE semantics.
For example, the coordinated verb phrase \emph{seizes and detains a victim} is transformed to  \emph{seizes a victim and detains a victim}.

\tabref{table:ace-fail} lists down a few English sentences that are not \ace-compliant, along with the reasons for their non-compliance. 
We then show corresponding sentences that are made \ace-compliant through syntactic transformation.
The transformed sentences become valid \ace sentences.
\tabref{table:ape-fail} lists down a few \ace sentences that are not \owl-compliant, along with the reasons for their non-compliance. 
We then show corresponding sentences that are made \ace-compliant through suitable transformation.
The transformed sentences can successfully be formalized into \owl.

\begin{table*}
    \centering
    \caption{Making English sentences \ace-compliant through syntactic transformation}
    \label{table:ace-fail}
    \begin{tabularx}{\linewidth}{|X|X|X|}  
        \hline
             \textbf{English Sentence}       & \textbf{Reason for non-compliance} & \textbf{Transformed Sentence} \\
        \hline
Every battery produces electricity. & Every noun should be prefixed by a determiner. & Every battery produces some electricity. \\
        \hline
An adenine is a purine base.	 & Multi-term expressions should be hyphenated. & An adenine is a purine-base.	 \\
        \hline
 An abdomen exists between thorax and pelvis. & Coordinated noun phrases have to agree in number and verb form. &  An abdomen exists between thorax and exists between pelvis. \\
        \hline
A kidnapper seizes and detains a victim. & Coordinated verb phrases have to agree in number and verb form. & A kidnapper seizes a victim and detains a victim. \\
        \hline
Every binomial consists of two terms. & Prepositional phrases should be attached to the corresponding verb. & Every binomial consists-of two terms. \\
        \hline
    \end{tabularx}
\end{table*}

\begin{table*}
    \centering
    \caption{Making \ace sentences \owl-compliant through syntactic transformation}
    \label{table:ape-fail}
    \begin{tabularx}{\linewidth}{|X|X|X|}  
        \hline
             \textbf{\ace Sentence}       & \textbf{Reason for non-compliance} & \textbf{Transformed Sentence} \\
        \hline
All kids play. & Intransitive verbs are not supported by \ape \owl generator. & All kids play something.\\
        \hline
Every person should learn some maths. & Modal verbs are not supported by \ape \owl generator. & Every person should-learn some maths.\\
        \hline
Every abacus efficiently performs some arithmetic. & Adverbs are not supported by \ape \owl generator. & Every abacus efficiently-performs some arithmetic\\
        \hline
A console houses some electronic instruments.  & Adjectives are not supported by \ape \owl generator. & A console houses some electronic-instruments.  \\
        \hline
    \end{tabularx}
\end{table*}

Note that it might be possible to apply syntactic transformation either by modifying the grammar rules of \ace or by improving \ape.
However, \ace is a formal language that was developed with a focus on knowledge representation.
Since the focus of our work is ontology authoring, we chose to have a separate module of syntactic transformation, which is independent of \aceape.
This module will be unaffected by any update which would be made to \aceape.
Hence, we decided to use \ace only as an intermediate language to generate axioms, instead of modifying \aceape.

\section{Ambiguity in Formalization}\label{section:ambiguity}

A natural language sentence can be ambiguous due to various reasons.
We note that there are different types of ambiguity such as
	lexical ambiguity, scope ambiguity, syntactic ambiguity and semantic ambiguity.
%
The concept of ambiguity is well-studied 
	in the context of NLP tasks such as POS tagging, 
	word sense disambiguation and sentence parsing\mycite{manning1999foundations}\mycite{jurafsky2000speech}.
For instance, syntactic parsing of a sentence can often lead to multiple parse trees.
Various algorithms are used to identify the most-probable parse tree.
However, as far as we know, there is no concrete study of ambiguity 
	in the context of formalization of sentences into \owl axioms.
%
We investigate the impact of two types of ambiguities, namely lexical ambiguity and semantic ambiguity.
In the following section, we describe each type in detail and how both are handled by the system.
%

\subsection{Lexical Ambiguity}\label{section:lex-amb}
Lexical ambiguity is a major type of ambiguity associated with formalization of sentences.
An ontology element (i.e. a class, relation or instance) can be lexicalized in more than one way due to various modeling possibilities.
It depends on factors such as the domain, individual preferences, and application. 
As a result, it is possible to generate multiple formalizations for the same sentence, by combining different lexicalizations of its elements in various ways. 
For example, consider the following sentence: 
\emph{Every vegetable pizza is a tasty pizza.}
Note the adjective \emph{tasty} that describes the noun \emph{pizza}.
While formalizing the sentence, this adjective can either be lexicalized as a separate concept or associated with the corresponding noun.
This lexical ambiguity with respect to the adjective results in two different formalizations, as shown below:
\begin{asparaenum}
	 \item \texttt{VegetablePizza} $\sqsubseteq$ \texttt{TastyPizza}
	 \item \texttt{VegetablePizza} $\sqsubseteq$ \texttt{Tasty} $\sqcap$ \texttt{Pizza}
\end{asparaenum}


Lexically ambiguous sentences can be disambiguated by using a point of reference such as an existing ontology\mycite{Emani2015}.
In that case, the lexicalization that best fits the reference can be chosen.
However, in the absence of suitable points of reference, the best possible way is to generate all possible lexicalizations.
Our system identifies lexical ambiguity associated with formalization
	and generates all possible formalizations of the sentence.
	
Williams\mycite{williams2013analysis} studies the syntax of identifiers from a corpus of over 500 ontologies. 
The identifiers chosen for the study are class identifiers, named individual identifiers, object property identifiers and data property identifiers.
According to the study, identifiers follow simple syntactic patterns and each type of identifier can be expressed through relatively few syntactic patterns.
These patterns are expressed using Penn POS tag set\mycite{Santorini1990}.
We adapt these syntactic patterns in our approach.

First, we POS-tag each word in the input text using Stanford part-of-speech tagger\mycite{Toutanova2003}.
Then, we use the aforementioned syntactic patterns to extract all possible identifiers from sentences.
By properly combining the identifiers, we generate all the formalizations of the sentence.
In case of the above example, the following phrases/words are extracted as identifiers: \emph{vegetable pizza}, \emph{is}, \emph{tasty pizza}, \emph{tasty} and \emph{pizza}. 
From these identifiers, our system generates both the given formalizations.

\subsection{Semantic Ambiguity}
Another type of ambiguity associated with formalization is semantic ambiguity. 
Due to semantic ambiguity, a sentence can have more than interpretation. 
For example, consider the following sentence:
\emph{Every driver drives a car.}
There are 3 possible interpretations of this sentence. 
The formalizations corresponding to each interpretation are:
\begin{asparaenum}
	\item \texttt{Driver} $\sqsubseteq$ $\exists$\texttt{drives.Car}
	\item \texttt{Driver} $\sqsubseteq$ $\forall$\texttt{drives.Car}
	\item \texttt{Driver} $\sqsubseteq$ $\exists$\texttt{drives.Car} $\sqcap$ $\forall$\texttt{drives.Car}
\end{asparaenum}
Note the quantifications associated with each axiom.
The first axiom denotes the correct interpretation (and hence the correct formalization) of the given sentence.
The existential quantification is appropriate due to the fact that a person should be driving \emph{at least} one vehicle to become a driver. 
A universal quantification is inappropriate due to the fact that a driver might drive \emph{any} vehicle, not necessarily a car.

Now consider a structurally similar sentence:
\emph{Every vegetable pizza is made of vegetable items.}
There are 3 possible interpretations of this sentence. 
The corresponding formalizations are:
\begin{asparaenum}
	\item \texttt{VegPizza}~$\sqsubseteq$~$\exists$\texttt{madeOf.VegItems}
	\item \texttt{VegPizza}~$\sqsubseteq$~$\forall$\texttt{madeOf.VegItems}
	\item \texttt{VegPizza}~$\sqsubseteq$~$\exists$\texttt{madeOf.VegItems}~$\sqcap$~$\forall$\texttt{madeOf.VegItems}
\end{asparaenum}
However, in this case, according to world knowledge and common-sense, 
	the correct interpretation is denoted by the third formalization.


It is necessary to have access to sufficient background knowledge to disambiguate such sentences, as demonstrated above.
However, in the absence of background knowledge,
	the best possible way is to generate possible multiple interpretations of the semantically ambiguous sentence.
We studied various sentence patterns and then manually identified the ones that are semantically ambiguous.
We also record as to how to map them to corresponding interpretations. 
In the online phase, the system checks whether the new sentence makes use of any one of the patterns that were identified beforehand.
In such cases, 
	the system generates all the corresponding interpretations of the sentence.
In the case of sentences that do not contain any of the patterns, 
	we generate only one interpretation of the sentence.
The list of sentence patterns that indicate ambiguity is extendable.

\section{Results and Discussion}\label{section:results}

The details of the evaluation of our framework are given below. All the datasets and evaluation results are available online\footnote{https://anon.to/SuLSzZ}.

\subsection{Datasets}
We use various types of datasets in our evaluation covering multiple domains.
This facilitates concrete evaluation of our approach and also ensures that our approach is neither dataset-oriented nor domain-dependent.
The datasets used for evaluation are \texttt{\pizza}, \texttt{\ssn}, \texttt{\vsao}, \texttt{\ppwiki}, \texttt{\ppwnet} and \texttt{\lexos}.
\texttt{\pizza}, \texttt{\ssn} and \texttt{\vsao} are obtained from Emani et al.\mycite{Emani2015}.
Each dataset consists of 25 sentences that are text descriptions.
Each dataset is based on a domain ontology. \texttt{\pizza} is based on Pizza ontology, \texttt{\ssn} is based on Semantic Sensor Network ontology and \texttt{\vsao} is based on Vertebrate Skeletal Anatomy ontology.
\texttt{\ppwiki} and \texttt{\ppwnet} are built from scratch.
We identified 40 classes from people-pets ontology\footnote{http://sadi-ontology.semanticscience.org/people-pets.owl}
and collected their descriptions from Wikipedia and WordNet resulting in  \texttt{\ppwiki} and \texttt{\ppwnet} datasets respectively.
\texttt{\lexos} is obtained from V{\"o}lker et al.\mycite{Volker2007a}.
This dataset contains 115 sentences covering a variety of domains.

\subsection{Axiom Generation}
We use the \ace parser, Attempto Parsing Engine (\ape), to translate \ace text into \owl axioms.
The target ontology language chosen for formalization is \owltwo, in OWL/XML syntax.

\begin{figure}
\resizebox{0.5\textwidth}{0.4\textheight}
{\begin{tikzpicture}
\tikzset{edge from parent/.style={draw,edge from parent path={(\tikzparentnode.south)-- +(0,-10pt)-| (\tikzchildnode)}}}
\tikzset{level 1/.style={level distance=40pt, sibling distance=-30pt}}
\Tree 
[.\node[label={above right:{ Every n:adenine is a n:purine-base and v:found-in a n:DNA.}}]{\nonterm{axiom}}; 
	[.\node[label={above right:{ Every n:adenine}}]{\nonterm{lexpr}};
		[.\node{\nonterm{\small CLASS}};
			[.\node{\leaf{Every adenine}};]
		]
    ] 
    [.\node{\nonterm{rexpr}};
			[.\node[label={above right:{a n:purine-base and v:found-in a n:DNA.}}]{\nonterm{intersection}};
				[.\node{\nonterm{clsExp}};
					[.\node[label={above right:{a n:purine-base}}]{\nonterm{classComb}}; 
						[.\node{\nonterm{\small CLASS}};
							[.\node{\leaf{a purine base}};]
						]
	              ]
				]
				[.\node{\nonterm{clsExp}};
					[.\node[label={above right:{ v:found-in a n:DNA}}]{\nonterm{existRes}};
						[.\node{\nonterm{\small PROPERTY}};
							[.\node{\leaf{found in}};]
						]
					[.\node{\nonterm{classComb}};
						[.\node[label={above right:{ a n:DNA}}]{\nonterm{\small CLASS}};
							[.\node{\leaf{dna}};]
						]
					]
				]
			]
			]
			]
		]
	]
]
\end{tikzpicture}}
\centering
\caption{Parse tree for the input sentence \emph{``Every adenine is a purine base found in dna."}}
\label{fig:parse-tree}
\end{figure}
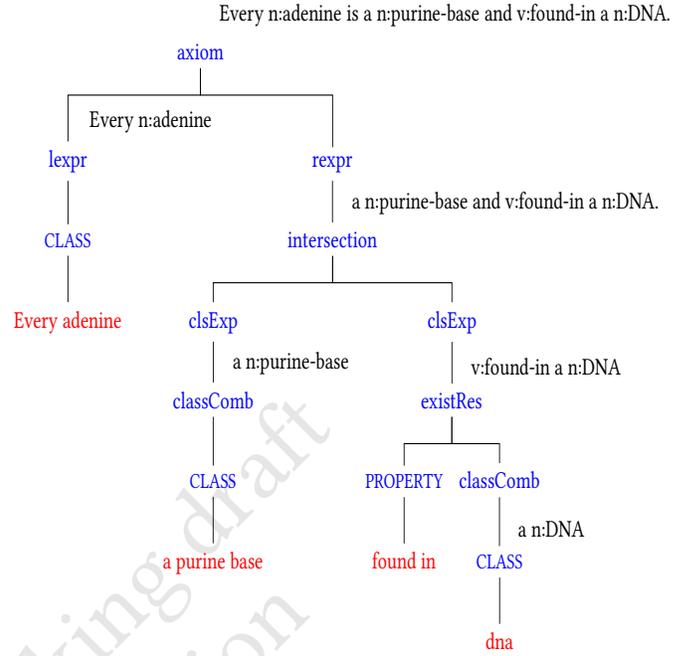

Few examples to demonstrate syntactic transformation and axiom generation
	are given in \tabref{table:syn-gen}.
The table presents sentences of various types and illustrates how the proposed system handles each type by generating the appropriate axiom.
Each example shows an input sentence and a formalization produced from the sentence by the system.
Here from the set of axioms produced by the system we only show the most appropriate one.
In \figref{fig:parse-tree} we present a parse tree for the input sentence of the first example \emph{Every adenine is a purine base found in DNA} .
The sentence is translated through syntactic transformation in a bottom-up fashion resulting in an equivalent \ace sentence.
The transformations performed on the sentence are shown at respective locations in the parse tree.
In this case, the resultant \ace sentence is shown at the root of the parse tree.
The corresponding OWL-DL axiom is \texttt{Adenine} $\sqsubseteq$ \texttt{PurineBase} $\sqcap$ $\exists$\texttt{foundIn.DNA}.


Few examples to demonstrate handling of ambiguity
	are given in \tabref{table:ambi-handle}. 
The given sentences are both lexically and semantically ambiguous.
Each example shows an input sentence and the various axioms produced from the sentence by the system.

\begin{table}
    \centering
    \begin{tabularx}{\linewidth}{r @{~} X}  
        \toprule 
             \textbf{Text}   & Every adenine is a purine base found in DNA.\\ 
             \textbf{Axm}  	& \texttt{Adenine} $\sqsubseteq$ \texttt{PurineBase} $\sqcap$ $\exists$\texttt{foundIn.DNA}\\
%


        \midrule 
             \textbf{Text}  & Sloppy giuseppe pizza is topped with mozzarella and parmesan.\\ 
             \textbf{Axm}  	& \texttt{SloppyGiuseppePizza} $\sqsubseteq$ $\exists$\texttt{toppedWith.Mozzarella} $\sqcap$ $\exists$\texttt{toppedWith.Parmesan}\\
%
%
%
        \midrule 
             \textbf{Text}  	& An interesting pizza is a pizza that has at least 3 toppings.\\ 
             \textbf{Axm}  	& \texttt{InterestingPizza} $\sqsubseteq$ \texttt{Pizza} $\sqcap$ $\geq$3\texttt{has.Toppings}\\

        \midrule 
             \textbf{Text}       	& Every abdication is the act of abdicating.\\ 
             \textbf{Axm}  	& \texttt{Abdication} $\sqsubseteq$ $\exists$\texttt{actOfAbdicating}.$\top$\\

        \midrule 
             \textbf{Text}       	& Every exotic species is a species that is not native to a region.\\ 
             \textbf{Axm}   	& \texttt{ExoticSpecies} $\sqsubseteq$ \texttt{Species} $\sqcap$ $\neg\exists$\texttt{isNativeToRegion}\\
             
	\bottomrule
    \end{tabularx}
    \caption{Syntactic transformation and axioms generation}
    \label{table:syn-gen}
\end{table}

\begin{table}
    \centering
    \begin{tabularx}{\linewidth}{r @{~} X}  
        \toprule
             \textbf{Text}  & Quarks possess color charge.\\ 
             \textbf{Axm}  		& \texttt{Quark} $\sqsubseteq$ \texttt{$\exists$possess.ColorCharge}\\
             \textbf{Axm}  		& \texttt{Quark} $\sqsubseteq$ \texttt{$\exists$possess.Color} $\sqcap$ \texttt{$\exists$possess.Charge}\\
             \textbf{Axm}  		& \texttt{Quark} $\sqcap$ \texttt{$\exists$possess.ColorCharge} $\sqsubseteq$ $\top$\\
             \textbf{Axm}  		& \texttt{Quark} $\sqcap$ \texttt{$\exists$possess.Color} $\sqcap$ \texttt{$\exists$possess.Charge} $\sqsubseteq$ $\top$\\

        \midrule
             \textbf{Text}  & A vegetarian pizza is an interesting pizza.\\ 
             \textbf{Axm}  	& \texttt{VegetarianPizza} $\sqsubseteq$ \texttt{InterestingPizza}\\
             \textbf{Axm}  	& \texttt{VegetarianPizza} $\sqsubseteq$ \texttt{Interesting} $\sqcap$ \texttt{Pizza}\\
             \textbf{Axm}  	& \texttt{VegetarianPizza} $\sqcap$ \texttt{InterestingPizza}  $\sqsubseteq$ $\top$\\
             \textbf{Axm}  	& \texttt{VegetarianPizza} $\sqcap$ \texttt{Interesting} $\sqcap$ \texttt{Pizza}  $\sqsubseteq$ $\top$\\

        \bottomrule
    \end{tabularx}
    \caption{Handling of ambiguity}
    \label{table:ambi-handle}
\end{table}

\subsection{Grammatical Coverage Analysis}
We evaluated the quality of \theGrammar by analyzing its grammatical coverage.
We compare the coverage of \ace and \theGrammar.
Our analysis shows that \theGrammar has larger grammatical coverage as compared to \ace.

\begin{table}
    \centering
    \begin{tabularx}{\linewidth}{|r|c|c|c|c|c|c|}
        \hline
	          					     \multicolumn{7}{|c|}{\textbf{DATASET}} \\
        \hline
	          					    & \textbf{\pizza} & \textbf{\ssn} & \textbf{\vsao} & \textbf{\ppwiki} & \textbf{\ppwnet} &\hspace{.15cm}\textbf{\lexos}\\
        \hline
	          \textbf{\inputsentences} & 25 & 25 & 25 & 40 & 40 & 115\\
        \hline
	          \textbf{\acesentences}   & 0 & 0 & 0 & 1 & 2 & 1\\
        \hline
	         \textbf{\tedeisentences} & 19 & 19 & 16 &7 & 22 & 22\\
        \hline
    \end{tabularx}
    \caption{Comparison of grammatical coverage of \theGrammar and \ace on various datasets 
    {\small (\textbf{IS} refers to the number of input sentences, \textbf{AS} refers to the number of \ace sentences, and \textbf{TS} refers to the number of \theGrammar sentences.)}
     }
    \label{table:coverage-analysis}
\end{table}

The analysis is reported, in detail, in \tabref{table:coverage-analysis}.
\textbf{IS} refers to the number of input sentences that are present in the dataset.
\textbf{AS} refers to the number of \ace sentences i.e. sentences that conform to \ace grammar.
\textbf{TS} refers to the number of \theGrammar sentences i.e. sentences that have at least one \theGrammar lexicalization.
Our system is guaranteed to generate at least one valid \owl axiom from every \theGrammar sentence.
For every dataset, the number of \theGrammar sentences is significantly more than the number of \ace sentences.
For instance, out of the 25 sentences in the \texttt{\pizza} dataset, none of the sentences are valid according to \ace, whereas 19 sentences are \theGrammar sentences. 
We can observe a similar pattern in every other dataset.

\subsection{Formalization}
We evaluated the number of formalizations generated by the proposed system.
We chose the datasets having at least 50\% \theGrammar sentences for this evaluation.
As can be observed from \tabref{table:coverage-analysis}, \texttt{\pizza}, \texttt{\ssn}, and \texttt{\vsao} have at least 50\% \theGrammar sentences.
Hence, these datasets are used for this evaluation.

The results are reported in \tabref{table:formalizations}. 
\textbf{\inputsentences} denotes the number of input sentences in the dataset.
\textbf{\totallexicalizations} denotes the total number of lexicalizations generated by lexical ambiguity handler for all the input sentences.
\textbf{\tedeilexicalizations} denotes the total number of valid \theGrammar lexicalizations i.e. lexicalizations that conform to \theGrammar.
\textbf{\interpretations} denotes the total number of interpretations generated by semantic ambiguity handler.
Each interpretation is the result of syntactic transformation of the \theGrammar lexicalizations.
\textbf{\validaceaxioms} denotes the total number of valid \ace axioms i.e. interpretations that conform to \ace.

As we can observe from the table, the proposed system generates many lexicalizations from a given input sentence, out of which many are valid \theGrammar sentences.
Based on various sentence patterns, we identify semantically ambiguous sentences, which have more than one interpretation. 
A major portion of the complete set of interpretations are valid \ace sentences, each of which can be converted to an equivalent \owl axiom.
For instance, from the 25 sentences in the \texttt{\pizza} dataset, 677 lexicalizations are generated, out of which 157 are valid according to \theGrammar.
Subsequently, 260 interpretations are generated, out of which 206 are valid according to \ace.

\def\pad{.66cm}


\begin{table}
    \centering
    \begin{tabular}{|@{\quad}@{\quad}r@{\quad}|@{\qquad}c@{\qquad}|@{\qquad}c@{\qquad}|@{\qquad}c@{\quad}|}
        \hline
	          					     \multicolumn{4}{|c|}{\textbf{DATASET}} \\
        \hline
	          					    & \textbf{\pizza} & \textbf{\ssn} & \textbf{\vsao}\\
        \hline
	          \textbf{\inputsentences} & 25 & 25 & 25\\
        \hline
	          \textbf{\totallexicalizations}   & 677 & 5964 & 560\\
        \hline
	          \textbf{\tedeilexicalizations}   & 157 & 297 & 93\\
        \hline
	          \textbf{\interpretations}   & 260 & 594 & 186\\
        \hline
	          \textbf{\validaceaxioms}   & 206 & 180 & 72\\
        \hline
    \end{tabular}
    \caption{Evaluation of formalization on various datasets
    {\small (\textbf{\inputsentences} denotes the number of input sentences, \textbf{\totallexicalizations} denotes the total number of lexicalizations, \textbf{\tedeilexicalizations} denotes the total number of valid \theGrammar lexicalizations, \textbf{\interpretations} denotes the total number of interpretations, and \textbf{\validaceaxioms} denotes the total number of valid \ace axioms)}
}
    \label{table:formalizations}
\end{table}

We also evaluated the correctness of the axioms.
We chose the \texttt{\pizza} dataset for this evaluation.
We manually authored gold standard set of axioms and compared it with axioms generated by the system. 
Out of the 25 sentences in the dataset, for 17 sentences, the human-authored axiom is indeed one of the alternatives given by the system.

\section{Conclusions and Future Works}
\label{section:conclusion}

In this paper, we propose an ontology authoring framework that extracts class expression axioms from natural language sentences through grammar-based syntactic transformation.
The input sentences that conform to the grammar are transformed into \ace.
We then use \ace parser to generate \owl axioms from the transformed text.
We also explore the effect of ambiguity on formalization and construct axioms corresponding to alternative formalizations of a sentence.

Our framework clearly outperforms \ace in terms of the number and types of sentences the system can handle.
In comparison with \lexo and other ontology learning systems, our framework is a robust way to generate ontological axioms from text.
\theGrammar helps in clearly defining the scope of the system and provides the ability to reject a sentence and ask for reformulation.
Employing \ace as an intermediate language aids formalization and reduces the complexity of the system.
The output of the system is compared against human-authored axioms and in a substantial number of cases, human-authored axiom is indeed one of the alternatives given by the system.

Our future works include enhancing the grammar so that the framework can handle a larger range of sentences and investigating the impact of other types of ambiguities such as scope ambiguity on formalization.

\bibliographystyle{ACM-Reference-Format}
\bibliography{library}

\end{document}